\documentclass[conference]{IEEEtran}
\IEEEoverridecommandlockouts
\usepackage{cite}
\usepackage{amsmath,amssymb,amsfonts}
\usepackage{algorithmic}
\usepackage{graphicx}
\usepackage{textcomp}
\usepackage{xcolor}
\usepackage{float}
\def\BibTeX{{\rm B\kern-.05em{\sc i\kern-.025em b}\kern-.08em
    T\kern-.1667em\lower.7ex\hbox{E}\kern-.125emX}}
\begin{document}

\title{Improving Mitosis Detection Via UNet-based Adversarial Domain Homogenizer\\
\thanks{* Indicates equal contribution.}
}

\author{\IEEEauthorblockN{1\textsuperscript{st} Tirupati Saketh Chandra*}
\IEEEauthorblockA{\textit{Electrical Engineering} \\
\textit{Indian Institute of Technology Bombay}\\
Mumbai, India \\
0000-0002-0325-5821}
\and
\IEEEauthorblockN{2\textsuperscript{nd} Sahar Almahfouz Nasser*}
\IEEEauthorblockA{\textit{Electrical Engineering} \\
\textit{Indian Institute of Technology Bombay}\\
Mumbai, India \\
0000-0002-5063-9211}
\and
\IEEEauthorblockN{3\textsuperscript{rd} Nikhil Cherian Kurian}
\IEEEauthorblockA{\textit{Electrical Engineering} \\
\textit{Indian Institute of Technology Bombay}\\
Mumbai, India \\
0000-0003-1713-0736}
\and
\IEEEauthorblockN{4\textsuperscript{th} Amit Sethi}
\IEEEauthorblockA{\textit{Electrical Engineering} \\
\textit{Indian Institute of Technology Bombay}\\
Mumbai, India \\
0000-0002-8634-1804}
}

\maketitle

\begin{abstract}
The effective localization of mitosis is a critical precursory task for deciding tumor prognosis and grade. Automated mitosis detection through deep learning-oriented image analysis often fails on unseen patient data due to inherent domain biases. This paper proposes a domain homogenizer for mitosis detection that attempts to alleviate domain differences in histology images via adversarial reconstruction of input images. The proposed homogenizer is based on a U-Net architecture and can effectively reduce domain differences commonly seen with histology imaging data. We demonstrate our domain homogenizer's effectiveness by observing the reduction in domain differences between the preprocessed images. Using this homogenizer, along with a subsequent retina-net object detector, we were able to outperform the baselines of the 2021 MIDOG challenge in terms of average precision of the detected mitotic figures.
\end{abstract}


\section{Introduction and Related Work}

The demand for robust models which perform well on unseen testing data has increased rationally in the last two years, especially in the medical domain. 

In many practical applications of machine learning models domain shift occurs after training, wherein the characteristics of the test data are different from the training data.  Particularly in the application of deep neural networks (DNNs) to pathology images, the test data may have different colors, stain concentration, and magnification compared to what the DNN was trained on due to changes in scanner, staining reagents, and sample preparation protocols. MIDOG2021 \cite{b20} was the first MICCAI challenge that addressed the problem of domain shift as it is one of the reasons behind the failure of machine learning models after training, including those for mitosis detection techniques when tested on data from a different domain (scanner) than the one used during the training. MIDOG2021 was the first MICCAI challenge that addressed the problem of domain shift due to change in scanner \cite{b20}.

 Domain generalization is the set of techniques that improve the prediction accuracy of machine learning models on data from new domains without assuming access to those data during training. Proposing and testing various domain generalization techniques was the main goal of the MIDOG2021 challenge. In this section, we introduce the most significant notable solutions for the MIDOG20201 challenge before introducing our proposed method.

 In \cite{b1} the authors modified RetinaNet network for mitosis detection \cite{b2} for mitosis detection by adding a domain classification head and a gradient reversal layer to encourage domain agnosticism. In this work, they used a pre-trained Resnet18 for the encoder. For their discriminator, it was a simple sequence of three convolutional blocks and a fully connected layer. The domain classifier was placed at the bottleneck of the encoder.
Breen et al \cite{b3} proposed a U-Net type architecture that outputs the probability map of the mitotic figures. These probabilities get converted into bounding boxes around the mitotic figures. They used a neural style transfer (NST) as a domain adaptation technique. This technique casts the style of one image on the content of another.
The method proposed by \cite{b4} consists of two parts, a patch selection and a style transfer module. To learn the styles of images from different scanners, they used a StarGAN.
A two steps domain-invariant mitotic detection method was proposed by \cite{b5}. This method is based on Fast RCNN \cite{b6}. For domain generalization purposes they used StainTools software \cite{b17} to augment the images. StainTools package decomposes the image into two matrices, a concentration matrix C and a stain matrix S. By combining the C and S matrices from different images they produced the augmented images.
A cascaded pipeline of a Mask RCNN \cite{b9} followed by a classification ensemble was proposed by \cite{b7} to detect mitotic candidates. A Cycle GAN \cite{b8} was used to transfer every scanner domain to every other scanner domain.
In \cite{b10} the authors used a stain normalization method proposed by \cite{b11} as a preprocessing step for the images.
Others like \cite{b12} merged Hard negative mining with immense data augmentation for domain generalization was proposed by \cite{b12}.
Stain normalization techniques such as \cite{b14} and \cite{b15} were used in \cite{b13} to account for the domain difference between images.
Almahfouz Nasser et al., \cite{b16} proposed an autoencoder trained adversarially on the sources of domain variations. This autoencoder makes the appearance of images uniform across different domains.

In this paper, we present our work which is an extension of our proposed method for MIDOG2021 \cite{b16}. 
Our contribution is three-fold and can be summarised in as follows. Firstly, we modified \cite{b16} by shifting the domain classifier from the latent space to the end of the autoencoder, which improved the results drastically. Secondly, we showed the importance of perceptual loss in preserving the semantic information which affects the final accuracy of the object detection part. Finally, unlike our previous work, training the auto-encoder along with the object detection network end-to-end improved the quality of the homogenized outputs substantially.

\section{Methodology}
\subsection{Notations}
\label{meth2:notations}
In domain generalization there are source (seen) domains which are shown to the model during training, and there are some target (unseen) domains which are used only during testing. Labelled samples from the source domains are represented by $D_{ls}$=$\{(x^{ls}_i,y^{ls}_i)\}^{N_{ls}}_{i=1}$, unlabelled source domains are represented by $D_{us}$=$\{(x^{us}_i)\}^{N_{us}}_{i=1}$ and labelled target domains are represented by $D_{lus}$=$\{(x^{lus}_i,y^{lus}_i)\}^{N_{lus}}_{i=1}$. Let the unlabelled images from all subsets be represented by $D_{all}$=$D_{ls}$ $\cup$ $D_{us}$ $\cup$ $D_{lus}$.

\subsection{Adversarial end-to-end trainable architecture}
\begin{figure}[H]
  \centering
    \includegraphics[width=0.45\textwidth]{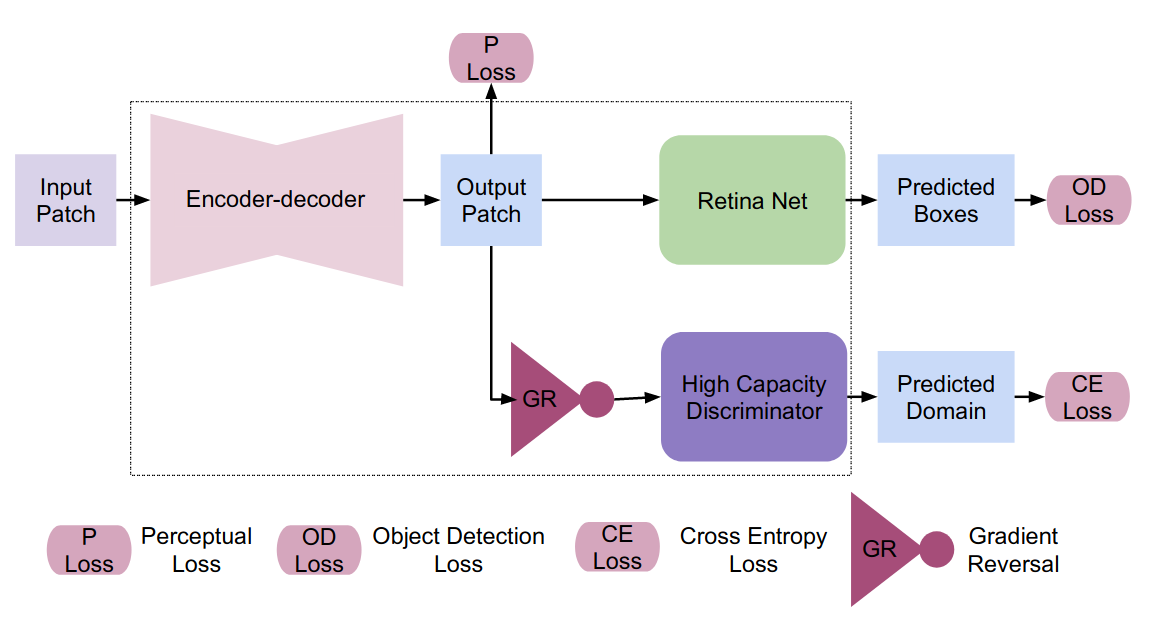}
    \caption[End-to-End trainable domain generalization architecture]{The pipeline of our proposed method for mitosis detection.} 
    \label{fig2:end-to-end} 
\end{figure}

Inspired by the work of \cite{b18}, we have used an encoder-decorder network to translate the patches from different domains (scanners) to a common space. The translated images are then passed through RetinaNnet \cite {b2} for object detection. The architecture also consists of an adversarial head with domain classification as an auxiliary task. This head encourages the encoder-decoder network to erase all the domain- specific information. The architecture of our method is as shown in figure \ref{fig2:end-to-end}.

\subsection{Training Objectives}
The object detection loss consists of bounding box loss ($\mathcal{L}_{bb}$) and instance classification loss ($\mathcal{L}_{inst}$). The bounding box loss ($\mathcal{L}_{bb}$) is computed as smooth L1 loss and the focal loss function is used for the instance classification ($\mathcal{L}_{inst}$).

In order to ensure that the images translated by encoder-decoder network contains the semantic information we have used a perceptual loss ($\mathcal{L}_{percp}$). We have used the perceptual loss based on pretrained VGG-16, which is proposed in \cite{b19}.

At the end of the adversarial head we have used standard cross entropy loss ($\mathcal{L}_{CE}$) for domain classification.

The overall loss for the end-to-end training is given by,
\begin{equation}
\mathcal{L}=\mathcal{L}_{bb}+\mathcal{L}_{inst}+ \lambda_{1}\mathcal{L}_{percep} + \lambda_{2}\mathcal{L}_{CE}
\label{loss:overall}
\end{equation}

\section{Data and Experiments}
\subsection{Dataset}
The experiments were conducted on MIDOG 2021 dataset \cite{b20} which consists of 50 whole slide images of breast cancer from four scanners namely Hamamatsu XR NanoZoomer 2.0, Hamamatsu S360, Aperio ScanScope CS2, and Leica GT450 forming four domains. Two classes of objects are to be detected namely mitotic figures and hard negatives. The whole slide images from scanners other than the Leica GT450 areis labelled. Small patches of size 512 x 512 are mined for supervised end-to-end training such that the cells belonging to at least one of the mitotic figures or hard negatives are present in the patch.

The seen and unseen domains i.e., the scanners are $D_{ls}$=\{Hamamatsu XR NanoZoomer 2.0, the Hamamatsu S360\}, $D_{us}$=\{Leica GT450\}, $D_{lus}$=\{Aperio ScanScope CS2\} (refer \ref{meth2:notations} for notations.)

\subsection{Implementation Details}

The model is implemented using Pytorch \cite{b21} library \cite{b21}. For supervised end-to-end training a batch size of 12 is used with equal number of patches being included from each scanner. Here the model is trained using FastAI \cite{b22} library default settings with an initial learning rate of $1e^{-4}$. In the equation \ref{loss:overall} we have set the values of hyperparameters as $\lambda_{1}$=10 and $\lambda_{2}$=25.


\subsection{Results}

Two classes of objects -- hard negatives, and mitotic figures -- are detected. The models are evaluated on $D_{ls} \cup D_{lus}$. Average precision at IoU threshold of 0.5 is used as metric for evaluation. End-to-end training with (AEC\_RetinaNet + Pecp) and without using perceptual loss (AEC\_RetinaNet) were tried. The results are compared with the reference algorithm DA\_RetinaNet \cite{b23}, RetinaNet \cite{,b24} with and without data augmentation. The results obtained are as shown in table \ref{table2:fin_results}.
\begin{table}[H]
\centering
\caption{Results obtained using end -to -end training of models}
\begin{tabular}{|l|l|l|l|} 
\hline
\textbf{Model} & \textbf{AP-Hard Neg} & \textbf{AP-Mitotic figures} & \textbf{mAP} \\ 
\hline
RetinaNet & 0.196 & 0.352 & 0.274 \\ 
\hline
RetinaNet +
  Aug & 0.238 & 0.619 & 0.429 \\ 
\hline
DA\_RetinaNet & 0.347 & \textbf{0.655} & 0.501 \\ 
\hline
AEC\_RetinaNet & 0.289 & 0.448 & 0.369 \\ 
\hline
AEC\_RetinaNet + Pecp & 0.248 & \textbf{0.72} & 0.484 \\
\hline
\end{tabular}
\label{table2:fin_results}
\end{table}

\begin{figure}[H]
  \centering
    \includegraphics[width=0.5\textwidth]{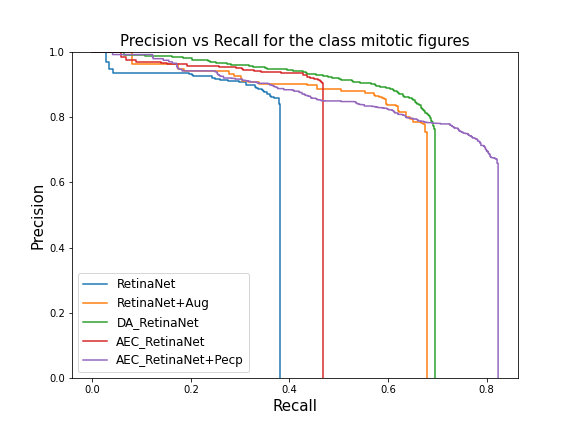}
    \caption[Precision vs. recall for the class mitotic figures]{The precision vs recall plot shows that the newly designed end-to-end model performs better than the baselines especially for the recall values.} 
    \label{fig2:PR plot} 
\end{figure}

The count of mitotic figures is an important clinical goal. So, the performance on the class of mitotic figures was our focused. The results in the above table show that the newly designed end-to-end training architectures performs better than the reference algorithm and the basic Retina-Net based algorithms. The best performance is in terms of recall metric, which is evident from the plot shown in figure \ref{fig2:PR plot}. 

The perceptual loss added at the output of the decoder helps in retaining the semantic information. This information which helps in better object detection. This is also validated by higher AP score obtained when perceptual loss component is added.

\begin{figure}[H]
  \centering
    \includegraphics[width=0.5\textwidth]{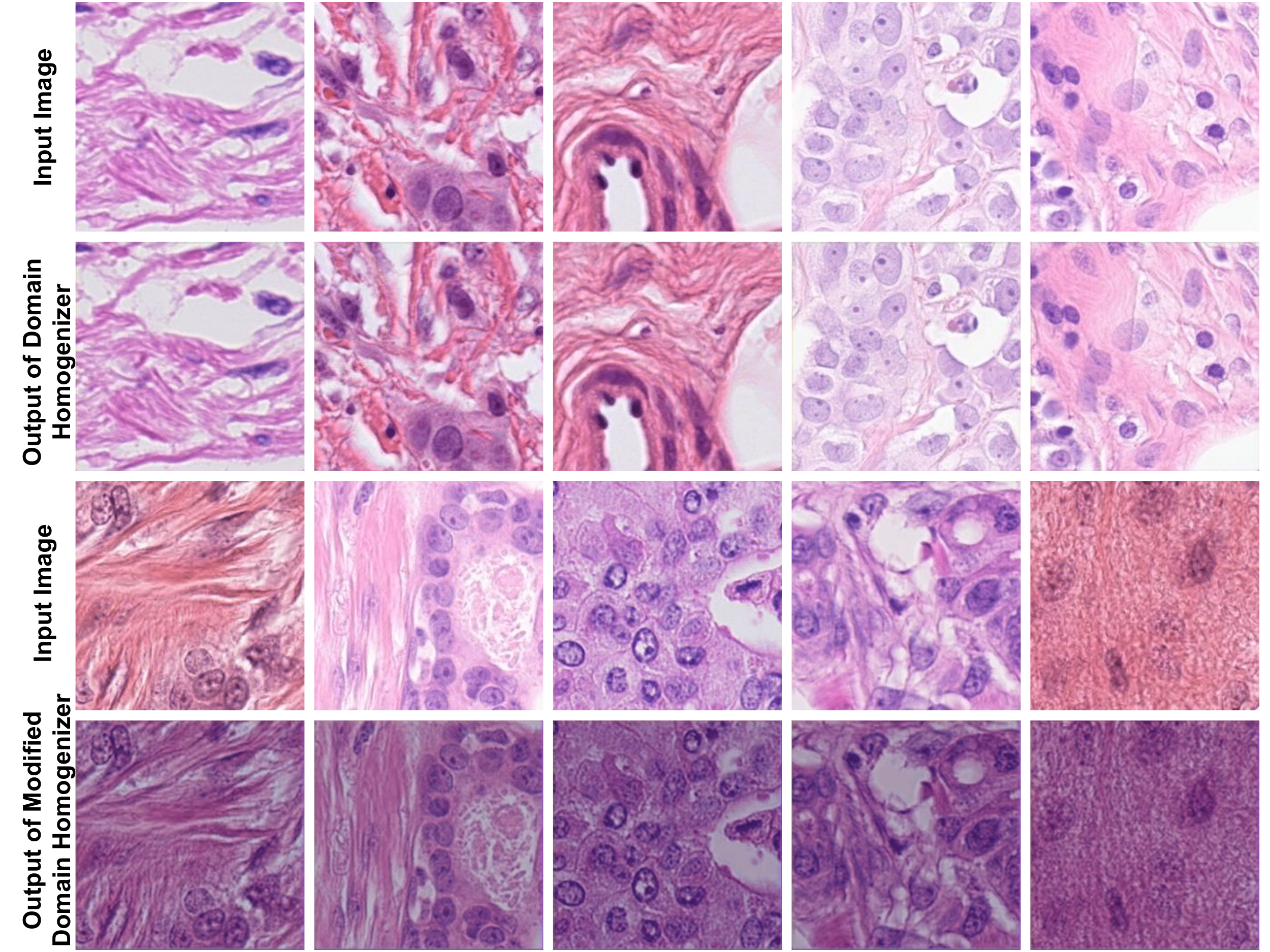}
    \caption{A visual comparison of the performances of the domain homogenizer and the modified domain homogenizer(proposed method).} 
    \label{fig3:results_vis} 
\end{figure}

As shown in figure \ref{fig3:results_vis} the modified domain homogenizer produced much more plausible images than the original domain homogenizer \cite{b16}.

\begin{figure}[H]
  \centering
    \includegraphics[width=0.5\textwidth]{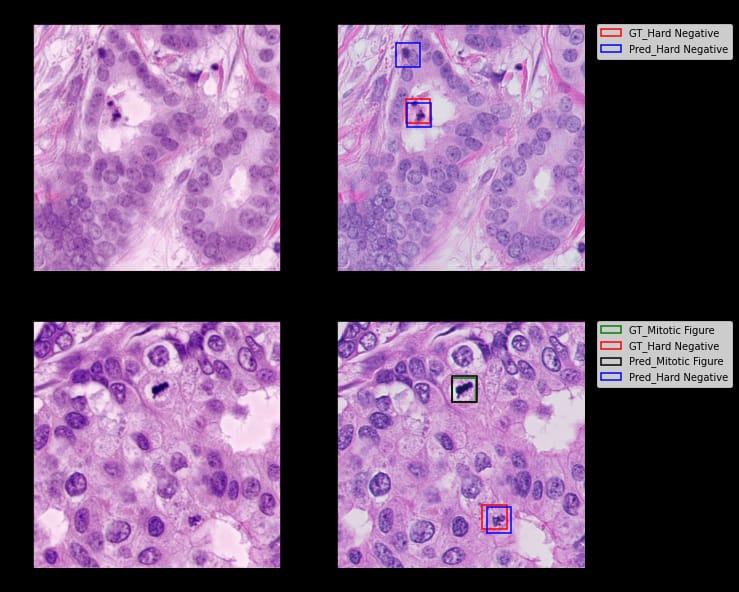}
    \caption{Two examples show the detection accuracy of our proposed method.} 
    \label{fig4:detection_res} 
\end{figure}

Besides, figure \ref{fig4:detection_res} shows the detection accuracy of our proposed method.

\section{Conclusions}

In this paper, we proposed a modified version of our previous domain homogenizer proposed by us and tested it on the data from for the MIDOG 2021 challenge 2021. We showed that the great impact of the position of the domain classifier has a significant impact on the performance of the homogenizer. [EXPLAIN HOW AND WHY] Additionally, our experiments revealed that training the homogenizer along with the object detection network end-to-end improves the detection accuracy by a significant margin. Finally, we showed that our method substantially improves upon the baseline of the MIDOG challenge in terms of mitotic figures detection. 

\vspace{12pt}

\end{document}